\documentclass{article}

\usepackage{arxiv}

\usepackage[utf8]{inputenc} 
\usepackage[T1]{fontenc}    
\usepackage{hyperref}       
\usepackage{url}            
\usepackage{booktabs}       
\usepackage{amsfonts}       
\usepackage{nicefrac}       
\usepackage{microtype}      
\usepackage{lipsum}
\usepackage{cite}
\usepackage{amsmath,amssymb,amsfonts}

\usepackage{multirow}
\usepackage{multicol}
\usepackage{graphicx}
\usepackage{textcomp}
\renewcommand\labelenumi{(\roman{enumi})}
\renewcommand\theenumi\labelenumi
\def\BibTeX{{\rm B\kern-.05em{\sc i\kern-.025em b}\kern-.08em
    T\kern-.1667em\lower.7ex\hbox{E}\kern-.125emX}}
    
\usepackage{enumitem}

\usepackage{algorithm}
\usepackage{algorithmic}
\usepackage{bm}
\usepackage{gensymb}
\usepackage{xcolor}
\usepackage{ulem}
\usepackage[all]{nowidow}
\usepackage{tablefootnote}
\usepackage{threeparttable}

\title{Partial Bandit and Semi-Bandit: Making the Most Out of Scarce Users' Feedback}

\author{
  Alexandre Letard \\
  Universit\'e d'Angers - LERIA\\
  Kara Technology - Dpt R\&D\\
  Angers, France \\
  \texttt{alexandre.letard@kara.technology} \\
   \And
 Tassadit Amghar \\
  Universit\'e d'Angers - LERIA\\
  Angers, France \\
  \texttt{tassadit.amghar@univ-angers.fr} \\
   \And
 Olivier Camp \\
  Groupe ESEO - ERIS\\
  Angers, France \\
  \texttt{olivier.camp@eseo.fr} \\
   \And
 Nicolas Gutowski \\
  Universit\'e d'Angers - LERIA\\
  Angers, France \\
  \texttt{nicolas.gutowski@univ-angers.fr} \\
}

\begin{document}
\maketitle

\begin{abstract}
Recent works on Multi-Armed Bandits (\textit{MAB}) and Combinatorial Multi-Armed Bandits (\textit{COM-MAB}) show good results on a global accuracy metric. This can be achieved, in the case of recommender systems, with personalization. However, with a combinatorial online learning approach, 
personalization implies a large amount of user feedbacks. Such feedbacks can be hard to acquire when users need to be directly and frequently solicited. For a number of fields of activities undergoing the digitization of their business, online learning is unavoidable. Thus, a number of approaches allowing implicit user feedback retrieval have been implemented. Nevertheless, this implicit feedback can be misleading or inefficient for the agent's learning. Herein, we propose a novel approach reducing the number of explicit feedbacks required by Combinatorial Multi Armed bandit (\textit{COM-MAB}) algorithms while providing similar levels of global accuracy and learning efficiency to classical competitive methods. In this paper we present a novel approach for considering user feedback and evaluate it using three distinct strategies. Despite a limited number of feedbacks returned by users (as low as 20 \% of the total), our approach obtains similar results to those of state of the art approaches.  
\end{abstract}

\keywords{Reinforcement Learning \and Combinatorial Multi-Armed Bandits \and User Feedbacks \and Recommendations Systems \and Learning Speed}

\section{Introduction}

Machine learning-based recommendation system are commonly used in various fields of activities \cite{GUT_2019_THESE}.
Among the methods used for recommendation, those relying on Multi-Armed Bandit (\textit{MAB}) approaches obtain interesting results in terms of global accuracy \cite{GUT_2019_THESE, DJA_2019_ARXIV}. 
This is more specifically the case with Combinatorial Multi-Armed Bandit (\textit{COM-MAB}) \cite{LOU_2016_THESE}. 
From an industrial perspective, some fields of activities, as sailing and yachting  \cite{WMU_2019_REP}, are initiating a digital transformation in order to provide such intelligent recommendation to their customers.
Issues relative to both smart homes and smart vehicles can be encountered in the field of mobile housing, to which seafaring belongs \cite{WMU_2019_REP}.
Indeed, there are multiple ways of using a recreational vehicle such as a boat : as a main or secondary residence, a means of transport or even as a way for pushing one's limits. 
Those scenarios depend on each user and handling context.
In recent years, several works have been carried out to promote the digitization of boating \cite{ADR_2003_ECML, AAR_2002_ECAI}.
However, those studies essentially deal with navigation automation, usually disregarding the other facets of such recreational vehicles.
Yet, all dimensions of the smart home/vehicle should be considered to offer more accurate recommendations to the sailors and improve their maritime experience. 
Ultimately, our intent is to design a smart boat, EVA\footnote{Entité de Voyages Automatisée - 
Automated Travel Entity}, whose behaviour will be guided by the users' needs and the way the boat is used. 
Our aim is to optimize energy consumption in order to minimize both the risk of energy shortage at sea and the environmental impact of the boat. 
We will implement Combinatorial Multi-Armed Bandit algorithms to dynamically identify amongst the services available aboard those that are essential to satisfy the users needs at any given moment.

In order to give customized recommendations, Combinatorial Multi-Armed Bandit approaches learn from the user feedbacks given at each iteration \cite{COM_2015_NIPS}. 
Known strategies for considering such feedbacks essentially rely on the observation of a full rewards vector for all given recommendations \cite{KAR_2016_NONE, COM_2015_NIPS}.
This vector can be hard to acquire, specifically when the rewards rely on explicit user feedbacks, e.g., a rating or an appreciation. 
In this article, we present a method based on Combinatorial Multi Armed Bandit techniques and aimed at mitigating the feedbacks issue in interactive systems. We experiment and evaluate our method in the field of yachting with the aim of implementing it on our smart boat EVA.

We propose a novel approach applying both \textit{Bandit} and \textit{Semi-Bandit} strategies on a variable size subset of recommended items.
We call our method "\textit{Partial Bandit with Semi-Bandit}" (\textit{P-BSB}).
We describe three strategies to determine the subset of recommended items that should be considered: \textit{Reinforce -  RE}, \textit{Optimal Exploration - OE} and \textit{ Randomized - RD}.

In our experiments, we apply a combinatorial approach to several single play Multi-Armed Bandits 
algorithms. 
We observe that the \textit{P-BSB} approach allows the use of Combinatorial Multi-Armed Bandit techniques with restricted feedbacks.
Also, we note that the global accuracy and the number of iterations needed to observe convergence (\textit{convergence time}) are congruent to those obtained by applying the classical \textit{Bandit} and \textit{Semi-Bandit} approaches for a given horizon greater than 10~000 iterations.

In summary, our contributions include:
1) The application of a combinatorial approach to several state of the art single play Multi-Armed Bandit algorithms, especially \textit{UCB2} which, to the best of our knowledge, has not been exploited in a combinatorial context;
2) The evaluation and comparison of those algorithms in terms of global accuracy and convergence time on several datasets obtained from real-world applications;
3) The formalization and proposal of three 
settings for a novel method for taking user feedbacks into account by Multi-Armed Bandits  -- "\textit{Partial Bandit with Semi-Bandit}" --. Our method aims at mitigating the issues relative to the retrieval of user feedbacks while preserving learning efficiency. 

This paper is organized as follows. Section~\ref{relatedwork} introduces the background on the selected techniques, reviews some related works and presents our motivation for this work. 
Section~\ref{ProblemaAndMethod} depicts our problem settings and outlines our method.
Section~\ref{Results} discusses our experimental evaluation.
Finally, we conclude and open up new perspectives of work in Section~\ref{Conclusion}.

\section{Background and Motivation}\label{relatedwork}

This paper is essentially related to Combinatorial Multi-Armed Bandit approaches, and more precisely to the strategies employed for taking the user feedbacks into account  in the learning process.
Thus, this section first introduces the Multi-Armed Bandit (\textit{MAB}) Problem \cite{ROB_1952} and the Contextual Multi-Armed Bandit (\textit{CMAB}) Problem  \cite{LI_2010_CTX}. Afterwards, we expose the key concepts of the combinatorial approach \cite{CHE_2013_ICML} and the strategies frequently used to consider the user feedbacks.
Lastly, we discuss some related works extracted from the literature and expose our motivations.

\subsection{The Multi-Armed Bandit Problem}\label{secMAB}

The Multi-Armed Bandit Problem has been widely investigated since its first formulation by Robbins in 1952 \cite{ROB_1952}.
As a result, there are nowadays numerous approaches under consideration \cite{GUT_2019_THESE}: stochastic \cite{AUER_2002}, non-stochastic \cite{AUER_2002_NONSTO} or Bayesian \cite{AGRAWAL_2012}. 
A \textit{MAB} problem includes a set $\mathcal{A} = \{a_1, ... a_m \}$ of \textit{m} independent arms, each arm $a \in \mathcal{A}$ being an item to recommend.
As part of a recommendation system, at each iteration $ t \in [1,\: T]$, with \textit{T} being a known horizon, a learning agent selects  an arm  $a_t \in \mathcal{A}$ according to its policy $\pi$ and recommends it to the user.
A part of the rewards vector $Y_t$\footnote{$Y_t$ is supposed to exist but can, actually, only be partially observed through non-omniscient approaches ($r_{t, a}$ in single play algorithms, $R_t$ in case of a combinatorial approach).}, associated to $\mathcal{A}$ is then uncovered to the agent which obtains an exclusive reward $r_{t, a} $ for the recommended item.
In this paper, we consider the case of Bernoulli bandits, where $r_{t,a} \in \{0,1\}$, with $r_{t,a}=0$ when the user does not validate a given recommendation and $r_{t,a}=1$ when the user is satisfied with the recommendation  \cite{GUT_2019_THESE}.   
In a stochastic setting, where the rewards are considered as being independent and identically distributed random variables across the arms, a \textit{MAB} algorithm aims at minimizing the cumulative regret $ \rho_T = T \mu^* - \sum_{t=1}^{T} r_{t, a} $, where $\mu^*$ is the reward expectation of the optimal arm, without prior knowledge about the rewards probabilities distribution $\mu_{a} \in [0,1]$ over each arm.
Minimizing regret is analogous to maximizing global accuracy $ Acc(T) = \frac{\sum_{t=1}^{T} r_{t, a}}{T}$, which is frequently used as an assessment criterion in the literature \cite{GUT_2019_THESE}.

In the specific case of Contextual Multi-Armed Bandit problems, the user context is employed to allow  customized recommendations.
This context is expressed as a vector $x \in \mathcal{X}$, $x \subseteq \mathbb{R}^d$, encoding the \textit{d} features of the user and their environment, e.g., profile (age, gender, occupation), preference, surroundings (localization, neighbourhood) or current activity.
In this variant, we suppose that there is a dependency between the reward expectation of each arm and the observed context.
In the case of a linear dependency, the expectation is expressed as follows: $\mathbb{E} [r_{t,a}|x_t] = \hat{\theta}^{\top}_{t, a} x_t$, where $\hat{\theta}_{t, a}$ is a coefficient vector associated to arm \textit{a}, initially null and estimated at each \textit{t} iteration.

\subsection{Combinatorial Multi-Armed Bandit}

The Combinatorial Multi-Armed Bandit problem can be seen as a generalization of the \textit{MAB} and \textit{CMAB} problems in which, at each iteration, a list of \textit{k}  arms, named "Super-Arm"  $S_t$, defined by $S_t = \cup_{i=1}^k a_i$, with $a_i = argmax^{\pi}_{a \in \mathcal{A} \text{\textbackslash{}} \{S_t\} } \mathbb{E}[R_{t,a}|x_t]$ is submitted to the user . 
One of the most popular combinatorial approaches is \textit{multiple play}.
It dynamically constructs $S_t$ by selecting $k$ arms \cite{CHE_2013_ICML, LOU_2016_THESE, COM_2015_NIPS} using one or several instances of a single play \textit{MAB} algorithm and aggregating them as $S_t$.
Thus, with this approach, each arm $a \in \mathcal{A}$ is considered individually by the learning process. 

Hence, the reward for $S_t$ can be expressed by $  S^{\top}_t R_t = \sum_{i=1}^k S_{t,i} \: R_{t,i}$, where $R_t $ is the $k$-size vector of observed rewards.
We name $\phi$ the strategy in place to take user feedbacks into account. It determines : 1) the composition of $R_t$ from $Y_t$ and $S_t$ ; 2) the agent's learning policy.

This extension of the \textit{MAB} and \textit{CMAB} problems has been frequently used on various fields of activities such as recommendation systems, finance or healthcare \cite{DJA_2019_ARXIV}.
Therefore, for our experiments, we choose to use algorithm~\ref{algo:banditTM} \cite{CHE_2013_ICML, LOU_2016_THESE, COM_2015_NIPS} which implements the multiple play approach.

In this paper, we also consider an overall reward, $r_{t}$ $\in \{0,1\}$ for $S_t$, where $r_t = 1$ if at least one of the suggested items in $S_t$ satisfies the user, 0 otherwise. Those rewards are then employed to determine the algorithm's global accuracy as defined in subsection~\ref{secMAB}. 

\subsection{Strategies for considering user feedbacks}

Various strategies have been proposed for \textit{COM-MAB} algorithms to take into account user feedbacks.
Most of them can be grouped in two main categories: \textbf{Bandit} \cite{ITO_2019_NIPS} and \textbf{Semi-Bandit} \cite{KAR_2016_NONE, COM_2015_NIPS} approaches.

With the \textit{Bandit} method, the agent only observes a cumulative reward, formulated as $R_{t,\: \phi_{B}} = S^{\top}_t R_t$, for a given Super-Arm $S_t$. That is, the individual reward of each arm in $S_t$ is not employed in the learning process. It is only used to compute the cumulative reward.
Under the \textit{Semi-Bandit} strategy however, the individual reward for each arm $a_{t,i}$ in $S_t$ is taken into consideration: $ R_{t,\: \phi_{SB}} = \cup_i \: S_{t,i} \: R_{t,i}$. 
In both cases, the entire rewards vector $R_t$, of size \textit{k} is essential for learning.

Actually, the cumulative reward required by the \textit{Bandit} strategy can be acquired externally, e.g., by directly asking the user for an overall rating. This way, the \textit{Bandit} approach can still be used if the full rewards vector $R_t$ is hard to acquire. However, one should know that, due to internal reasons e.g., imperfect or subjective approximations, limitations on the possible types of feedback acquired, user experience issues, or other external factors affecting the user's frame of mind, user's feedback is often biased \cite{TAN_2019_AAMAS}. This can make the best arms identification harder and, as such, lead to inadequate recommendations. Since an overall evaluation cannot express the user's opinion on every suggested items, this setting of the \textit{Bandit} strategy makes this issue even stronger. Moreover, by discarding the underlying arms information (their individual rewards), the algorithm's learning may be less efficient \cite{CHE_2013_ICML}. For those reasons, and because biaised feedbacks cannot be modelled, we consider in this paper the \textit{Bandit} strategy with an internally computed cumulative reward.


In the most recent works, the \textit{Semi-Bandit} approach seems predominant \cite{KAR_2016_NONE}.
This method has been adapted to fit the needs of different real-world application scenarios, e.g., the cascading bandit model for which the user feedback is expressed by a click on one of $S_t$'s items, and where the position of the clicked item is then used to infer the rewards associated to the others arms \cite{SHU_2016_ICML}.  

\subsection{Partial Methods}

A \textit{partial} user feedbacks consideration strategy is an approach where the observed rewards vector $ R_t$ is only defined for a subset $P_t \subseteq S_t$ of the Super-Arm $S_t$.
Formally, for a non-partial approach, the size of the observed rewards vector is expressed by $|R_t| = |S_t|$, while with a partial method this size is formulated as $|R_t| = l$ with $l < |S_t|$.   
In the current state of the art \cite{GRA_2017_ARXIV, LUE_2016_ARXIV, SAH_2019_NIPS}, the size $l$ of the observed rewards vector is considered constant. 

Hence, Grant et al. \cite{GRA_2017_ARXIV} use a partial \textit{Semi-Bandit} approach based on filtering $R_t$ by applying a binomial distribution. 
Luedtke et al. \cite{LUE_2016_ARXIV} also employ a partial \textit{Semi-Bandit} method under which a subset of $S_t$ is uniformly selected among all the subsets with size \textit{l} of $S_t$.
Similarly, Saha and Gopalan \cite{SAH_2019_NIPS} provide a partial \textit{Semi-Bandit} strategy considering the feedbacks on the top \textit{l}-ranking-arms as well as proofs of its regret lower bounds. 

The \textit{Partial Bandit with Semi-Bandit} (\textit{P-BSB}),  strategy that we propose differs from existing strategies and considers subsets of variable size.
From an industrial point of view, at each iteration, our goal is to use the maximum number of feedbacks a user agrees to give without exceeding it.
Another difference with other methods, apart from \cite{SAH_2019_NIPS}, is that with \textit{P-BSB}, the construction of the rewards vector $R_t$ is based either on a random selection (setting \textit{RD}) or on the user feedbacks observed up to iteration $t-1$ (settings \textit{RE} and \textit{OE}).
Finally, \textit{P-BSB} differs from other existing approaches by applying both rewards (\textit{Bandit}: $R_{t, B}$ and \textit{Semi-Bandit}: $R_{t, SB}$), computed from the partial user feedbacks vector (corresponding to $P_t$'s arms' rewards), when a recommended arm has indeed given satisfaction to the user.

\begin{figure}[ht]
\centering
\begin{minipage}{.7\linewidth}
\begin{algorithm}[H]
\caption{Multiple Play Bandit}\label{algo:banditTM}
\begin{algorithmic}[1]

\REQUIRE
Instance of a single play bandit and its specific settings. \\

$\pi$: Policy. \newline
$\mathcal{A}$: Set of available arms. \newline
$\textit{k}$: Number of items to recommend at each iteration. \newline
$Y_t$: Real Rewards vector.\newline
$\textit{T}$ : Horizon. \newline
$x \in \mathcal{X} $ : User context.\newline
$\phi(S_t, Y_t)$ : User feedbacks consideration strategy.\medbreak

\ENSURE Initialize the bandit instance according to $\pi$
          \medbreak

\FOR{$t = 1$ \TO \textit{T}}

    \STATE Consider $x_t \in X$: a user $u$ and their context \label{selecX}
        
    \STATE $S_t \gets \emptyset$ \medbreak
        
    \FOR{$i = 1$ \TO \textit{k}}
        
        \STATE Select item $a_i \in \mathcal{A}$\textbackslash{} \{$S_t$\} accordingly to $x_t$ and $\pi$
            
        \STATE $S_t \gets  S_t \cup \: a_i$
    \ENDFOR \medbreak

    \STATE Recommend $S_t$ to  user $u$ 
    \STATE Receive the overall reward $r_{t}$ for the recommendation $S_t, r_{t} \in \{0,1\}$
    \STATE Compute $R_t$ from $Y_t$ and $S_t$ according to $\phi$\label{determinerRt}
    \STATE Update policy $\pi$ with $R_t$ according to $\pi$ and $\phi$\label{MAJPI} \medbreak

\ENDFOR

\end{algorithmic}
\end{algorithm}
\end{minipage}
\end{figure}

\subsection{Motivation}

The aim of \textit{MAB} and \textit{CMAB} algorithms is to maximize their global accuracy \cite{GUT_2019_THESE}.
Therefore, user feedbacks and how they are considered by the algorithm play a crucial role in the agent's learning process.
However, the retrieval of a complete rewards vector $R_t$, which is fundamental for the \textit{Bandit} and \textit{Semi-Bandit} approaches, can be hard or even impossible to achieve in some real-world applications.

Numerous fields of activities, like seafaring, are just beginning their digital transformation and, as such, do not have any useful datasets for the offline learning of an agent. 
Nevertheless, if the system's recommendations do not meet the users' needs and expectations they are likely to turn away from the application.
For this reason, such a recommendation system should ensure  fast online learning.
Hence, we argue that the number of iterations needed to observe the convergence of the system's global accuracy (the \textit{convergence time}), which is representative of the learning speed of the agent, must be taken into account as an assessment criterion for the considered \textit{COM-MAB} and \textit{COM-CMAB} algorithms.

Also, even the most recent extensions of the \textit{Semi-Bandit} and \textit{Bandit} approaches, such as the cascading model \cite{SHU_2016_ICML}, can be troublesome or inadequate to use if the user feedbacks need to be explicitly retrieved, e.g., as a rating to identify the key points amongst a set of activities and stops on a cruise defined and suggested by the agent. In such a setting, the activities and stops would be the arms of a \textit{COM-MAB} / \textit{COM-CMAB} algorithm.
The potentially large number of feedbacks directly requested from the user, required to respond to such a scenario, could also divert the users from the application.

Therefore, the \textit{Partial Bandit with Semi-Bandit}  approach, introduced in this paper, is firstly built on the prior identification of a subset $P_t$ of $R_t$, of cardinality $\psi_t$ varying according to the number of feedbacks the user agrees to give. 
Then, \textbf{P-BSB} uses a \textit{Bandit}-like consideration on $S_t$ together with a \textit{Semi-Bandit} like approach on $P_t$. 
This twin reward awarded to some arms aims at increasing the agent's learning speed while favoring the arms which have genuinely given satisfaction to the user.
Through this novel method, our intent is to simplify the use of the \textit{COM-MAB} / \textit{COM-CMAB} algorithms on a wider scope of real-world applications by : 1) reducing the number of feedbacks requested from the user ; 2) reaching similar performances to those obtained by traditional approaches.

\section{Problem Setting}\label{ProblemaAndMethod}

In this section we state our problem and describe our method: \textit{Partial Bandit with Semi-bandit (P-BSB)}.
Our approach specifically focuses on the exploitation of user feedback and relies on the combination of \textit{Bandit} and \textit{Semi-Bandit} strategies, which have been extensively studied in the literature.
Herein, we apply this combination to a restricted number of user feedback observations.


\subsection{Problem Statement}

Let $\mathcal{X} \subseteq \{0,1\}^d$ be the set of \textit{d}-dimensional context vectors that characterize users and their environments, e.g., at each iteration $t$ of a given finite horizon $T$, $x \in \mathcal{X}$ is a binary vector encoding features of sequential arriving users with their contexts, pending for recommendations like: special scheduled tourism activities of the day, venues to visit, spots to explore.
In non contextual cases, i.e., without any context to rely on (context-free dataset) or when operating with algorithms using a context-blind policy, we set $x_t$ as a plain identifier, namely: $\forall t\in T, x_t=\vec{0}$.

Let $\mathcal{A}=\{a_1, ..., a_m\}$ denote the set of available items to be recommended by a \textit{COM-MAB} / \textit{COM-CMAB} algorithm according to a given policy $\pi$ and $\mu = \{\mu_{1},..., \mu_{m}  \}$ the rewards expectation distribution of each arm $a \in  \mathcal{A}$. 
At each iteration $t \in [1,T]$, let $S_t \subseteq \mathcal{A}$ be a subset of size $k<m$ of items built from $\mathcal{A}$, namely, \textit{super arm} $S_t$.
At each time point $t$, according to $\pi$ and $\mu_t$, the optimal super arm $S_t$ should be recommended to arriving user $u_t$ represented by a  context vector $x_t$. 
Then, let $r_{t} \in \{0,1\}$ be a global reward associated with super arm $S_t$, used to compute the agent's global accuracy $Acc^{\pi}(T)$, as defined in subsection~\ref{secMAB}. Let $Y_t$ denote the real rewards vector associated with each arms $a \in \mathcal{A}$. Finally, let $R_t \subseteq Y_t$ denote the rewards vector which can be actually observed by the agent and determined from the user's feedback.
\textit{Bandit} and \textit{Semi-Bandit} strategies assume that $R_t = Y_t$ for the \textit{k} arms encompassed by $S_t$. 
However, in several real-world applications, since one cannot observe rewards without obtaining explicit user feedbacks for the whole set of arms of $S_t$, this requirement thus becomes hard to meet.

Since overloading users with too many request for feedback can be detrimental or impossible in many real-world applications, such considerations led us to propose a novel approach aiming at decreasing the number of sollicitations for user feedback while ensuring an efficient learning.

In the next subsection, we introduce our novel approach built upon the combination of \textit{Bandit} and \textit{Semi-Bandit} approaches operating with a partial setting on a restricted subset of $S_t$.


\subsection{Partial Bandit with Semi-Bandit: P-BSB}

Herein, we focus on specific real-world applications where users must explicitly give their feedback and where $|R_t| = \psi_t \leq k$ is thus a stochastic variable representing the user's capacity to give a feedback (i.e., depending on e.g., their availability, their interest, their mood).
In this article, this capacity is referred to as $\psi$, the user's "\textit{patience}".

Similarly as for classical approaches, \textit{P-BSB} aims at building $R_t$ and using it in the agent learning process (See lines~\ref{determinerRt}~and~\ref{MAJPI} of Algorithm~\ref{algo:banditTM}).
The first step of \textit{P-BSB} is to determine a subset $P_t \subseteq \mathcal{S}_t$ of cardinality $\psi_t$ for which rewards will be observed by the agent such that $P_t = \cup_{i=1}^{\psi_t} a_i$.

Thus, in the first step of \textit{P-BSB}, we propose three different settings to determine which arms $a_i$ to consider: \smallbreak
\begin{itemize}
    \item \textbf{Reinforce - RE}: selects the $\psi_t$ arms from $S_t$ 
    having the highest reward 
    expectation $\mathbb{E}[R_{t,a} |x_t]$, 
    namely: \smallbreak
    \begin{equation}\label{(1)}
    a_i = argmax^{\pi}_{a \in S_t \text{ \textrm{\textbackslash{}}} \{P_t\} } \mathbb{E}[R_{t,a}|x_t]
    \end{equation}
    \medbreak
    \item \textbf{Optimal-Exploration - OE}: 
    selects the $\psi_t$ arms from $S_t$ from which associated rewards values have been 
    the least frequently observed at iteration \textit{t}, namely: \smallbreak
    \begin{equation}\label{(2)}
    a_i = argmin_{a \in S_t \text{ \textrm{\textbackslash{}}} \{P_t\} }~obs_{a,t} 
    \end{equation}
    \medbreak
    where $obs_{a,t}$ is the number of times the associated reward to arm $a_i$  has been observed up to iteration $t$.\medbreak
    \item \textbf{Randomized - RD}: 
    selects $\psi_t$ arms from $S_t$ randomly, namely: \smallbreak
     \begin{equation}\label{(3)}
    a_i = random(S_{t},\psi_t)
    \end{equation} \medbreak
    where $random(S_{t,\psi_t})$ selects $\psi_t$ distinct arms randomly from $S_{t}$.
    \smallbreak
\end{itemize}

The second step of \textit{P-BSB} operates in a similar way for all the three settings and consists in computing the rewards vector $R_t$ from $Y_t$ for each of the $\psi_t$ arms, such that:
\[ R_t = \cup_{i \in P_t} Y_{t,i}\]

The third step of \textit{P-BSB} (common to all settings again) combines a \textit{Bandit} kind strategy with a \textit{Semi-Bandit} kind strategy on each arm in $P_t$ \footnote{Then, if $a \in P_t$, two rewards are observed by the agent on arm $a$: $R_{t, B}$ and $R_{t, SB_a}$. If $\psi_t=0$, No reward are observed by the agent at iteration \textit{t}.} :

\[\forall a \in S_t , r_{t, a} = r_{t-1, a} + R_{t, B}\] and if $a \in P_t$ then $r_{t, a} = r_{t, a} + R_{{t, SB}_a}$ \smallbreak 

  where $r_{t,a}$ is the sum of rewards observed for arm \textit{a} up to iteration \textit{t}, and :

    \[R_{t, B} =  P_t^{\top}\: R_t\]
and $\forall i \in P_t$:
    \[ R_{t, SB} = \cup_i \: P_{t,i} \: R_{t,i}\] 
    \smallbreak
Note that in the \textit{Bandit} kind strategy, the cumulative reward $R_{t, B}$ is computed from the $\psi_t$ observed rewards, on each \textit{k} arms of $S_t$. 
This article sheds light on the \textit{RE} setting in algorithm~\ref{algo:RE}.
The two other settings (\textit{OE} and \textit{RD}) are similar except for their strategy to compute $P_t$.
Thus, to use these settings with algorithm~\ref{algo:RE}, we must replace line~\ref{lineAlgStrat}   with the policy of equation~\ref{(2)} to apply \textit{OE} or with the policy of equation~\ref{(3)} to apply \textit{RD}.

\textbf{RE} aims at rapidly reaching an exploitation step by favoring the selections of optimist actions.
On the other hand, \textbf{OE} follows a full exploration strategy aiming at strengthening the knowledge the agent has on the rewards expectation distribution ($\mu_1,..., \mu_k$).
Lastly, \textbf{RD} uses a random selection strategy and thus confers a compromise between exploitation and exploration. 

\section{Empirical Evaluation}\label{Results}

This section describes the empirical offline evaluation of the method we propose. 
This evaluation step is the preliminary and mandatory groundwork to be validated before we can integrate our approach into our marine recommendation system.
Hence, in this section we first introduce the datasets and algorithms we use to evaluate \textit{P-BSB}.
Then, we expose our experimental protocol, and
finally, we discuss the results we obtain.

\begin{figure}[ht]
\centering
\begin{minipage}{.7\linewidth}
\begin{algorithm}[H]
\caption{\textit{P-BSB} - RE}\label{algo:RE}
\begin{algorithmic}[1]

\REQUIRE

$S_t$, the super arm recommended to user. \newline
$Y_t$, the real rewards vector.\newline
$\pi$, the agent policy.\newline
$\psi_t$, the number of rewards that can be observed.\medbreak

\WHILE{ $|P_t|  <  \psi_t$ }

    \STATE Construct $P_t$ with  \newline
    $P_{t} = P_{t} \cup \: argmax^{\pi}_{a \in S_t \text{\textbackslash{}} \{ P_t\} } \mathbb{E}[R_{t,a}|x_t]$\label{lineAlgStrat} \newline
    (according to Equation~\ref{(1)})

\ENDWHILE \medbreak

\FOR{$i \in P_t $ }
        
    \STATE Construct $R_t$ with $R_{t} =  \cup_i \: Y_{t,i}$
    \STATE Apply \textit{Semi-Bandit} strategy to $R_t$: $ R_{t, SB} = \cup_i \: P_{t,i} \: R_{t,i}$
        
\ENDFOR \medbreak

\STATE Apply \textit{Bandit} strategy to $R_t$: $R_{t, B} =  P_t^{\top} \: R_t$\medbreak

\FOR{ $a \in S_t $}
        
    \STATE Update policy $\pi$ with $r_{t, a} = r_{t-1, a} + R_{t, B}$\medbreak
    
    \IF{$ a \in P_t$}
    
        \STATE Update policy $\pi$ with $r_{t, a} = r_{t, a} + R_{{t, SB}_a}$
        
    \ENDIF\medbreak
    
\ENDFOR \medbreak

\end{algorithmic}
\end{algorithm}
\end{minipage}
\end{figure}


\subsection{Datasets}

\footnotetext[4]{https://archive.ics.uci.edu/ml/datasets/covertype}
\footnotetext[5]{https://archive.ics.uci.edu/ml/datasets/Poker+Hand}
\footnotetext[6]{https://www.kaggle.com/assopavic/recommendation-system-for-angers-smart-city}
\footnotetext[7]{https://www.kaggle.com/vikashrajluhaniwal/jester-17m-jokes-ratings-dataset}
\footnotetext[8]{https://grouplens.org/datasets/movielens/100k/}

Datasets have been chosen according to different criteria in terms of scale: Number of instances (from 942 to 1 025 010),
number of features (from 0 to 54), and number of arms (from
7 to 1682).
The evaluation of our proposal is based on five real world datasets (see Table~\ref{table:dataset}):
\begin{itemize}
    \item \textbf{Covertype}\footnotemark[4] and \textbf{Poker Hand}\footnotemark[5] 
    are datasets to scale-up our experiment both in terms of number of instances and number of features;
    \item \textbf{RS-ASM}\footnotemark[6] is a services recommendation dataset focused on smart cities \cite{GUT_2019_THESE};
    \item \textbf{Jester}\footnotemark[7] is a non-contextual joke recommendation dataset;
    \item \textbf{MoviesLens}\footnotemark[8] is a movie recommendation dataset which has been used extensively in the literature.\smallbreak
    
\end{itemize}

Note that \textbf{Jester} and \textbf{MoviesLens} are datasets to scale-up our experiment in terms of number of arms and where items to recommend are rated on a scale from 0 to 5.
In our experiments,  we will consider that reward $R_{t,a}=1$ if the rating is greater than or equal to $4$, and $R_{t,a}=0$, otherwise.

\begin{table}
\renewcommand{\tablename}{Table}
\begin{center}
\resizebox{0.5\textwidth}{!}{
\begin{tabular}{cccc} 
  \hline
  Datasets & Instances & Arms & Features \\
  \hline
  Covertype\footnotemark[4] & 581 012 & 7 & 54 \\
  Poker\footnotemark[5] & 1 025 010 & 10 & 10 \\
  RS-ASM\footnotemark[6] & 2 152  & 18 & 50 \\
  Jester\footnotemark[7] & 59 132 & 150 & 0 \\
  MovieLens\footnotemark[8] & 942 & 1682 & 23 \\
  \hline
\end{tabular}
}
\end{center} 
\caption{Datasets}\label{table:dataset}
\end{table}

\subsection{Algorithms}
The method we evaluate focuses on how user feedback is considered.
Thus, its strategy operates independently from the policy $\pi$ of the agent, i.e., independently from the chosen \textit{COM-MAB} / \textit{COM-CMAB} algorithm.
We run Algorithm~\ref{algo:banditTM} with the most popular single play bandits algorithms to evaluate their global accuracy ($Acc(T)$) and convergence time ($t_c$).
Hence, in this paper, regarding previous evaluations of both \textit{COM-MAB} / \textit{COM-CMAB} \cite{CHE_2013_ICML, LOU_2016_THESE, CHEN_2018_NIPS}, we consider the following popular algorithms: 
\begin{itemize}
    \item \textbf{MAB}: $\varepsilon$-\textit{greedy} \cite{SUTTON_1998}
    with $\varepsilon=0.0009$, \textit{Thompson Sampling (TS)} \cite{AGRAWAL_2012}, \textit{UCB} \cite{AUER_2002} and \textit{UCB2} \cite{AUER_2002_FINITE};\smallbreak
    \item \textbf{CMAB}: \textit{LinUCB} \cite{LI_2010_CTX} and \textit{LinTS} \cite{AGRAWAL_2013}.
\end{itemize}

\subsection{Experimental Protocol}
To simulate a data stream of arriving users and contexts (see line~\ref{selecX} of algorithm~\ref{algo:banditTM}), we randomly select them one by one from the whole dataset. For each dataset, we scale a finite time horizon $T=10~000$.
In this article we introduce the concept of convergence time $t_c$, which refers to the time point wherefrom the measured global accuracy $Acc(t_c)$ is in a range of $\pm \delta$ (in this work we chose $\delta = 1 \%$) around the final global accuracy $Acc(T)$ (see subsection~\ref{secMAB}) : \smallbreak

$\forall t \geq t_c$: 
\begin{center}
$Acc(T)- \delta \leq Acc(t) \leq Acc(T) + \delta$, with \(\delta=0.01\)\smallbreak
\end{center}

For each experiment and for each dataset, algorithms recommend a 3-tuple of items ($k=3$) at each iteration $t$.
Herein, we compare our approaches \textbf{RE}, \textbf{OE} and \textbf{RD} with both competing strategies: \textbf{bandit} and \textbf{semi-bandit}.
For each  setting of \textbf{P-BSB}, the user "\textit{patience}" $\psi_t$ is determined at each iteration by a random variable ranging from $0$ to \textit{k}. 

In order to recommend a 10-tuple of items ($k=10$) using the \textbf{Jester} and \textbf{MoviesLens} datasets, one follows a similar process with $\psi_t$ from 0 to 4. Note that both datasets are remarkable in their large number of arms and thus allow to experiment with values of $ \psi_t  \ll k \ll m$.

Then, in each experimental case and for each algorithm, we simulate 10 cyclical iterations of $10~000$ rounds.

In Table~\ref{table:Jester} and Table~\ref{table:Movies} we observe each algorithm's global accuracy and convergence time: Their average and standard deviation with $k=10$ and $0 \leq \psi_t  \leq 4$.
This case is particularly relevant since, at each  \textit{t} iteration, the number of non-observed rewards under a \textit{P-BSB} strategy is higher. Those experiments are therefore more representative of the results we expect to obtain in our final application. 

In the next subsection we analyse and discuss the results we obtain. 

\subsection{Results Analysis}

\begin{table}[ht]
  \centering
\renewcommand{\tablename}{Table}
\resizebox{0.5\textwidth}{!}{
\begin{tabular}{|c|c|c|c|} 
\cline{1- 4} 
\textbf{Algorithm}	&\textbf{Strategy}	&\textbf{$Acc(T)$} &\textbf{$t_c$}		 \\ \hline\hline  

\multirow{5}{*}{$\varepsilon$-greedy}	&Bandit	            &$0,833$ {\tiny $\pm 0,002$}	&$1461$ {\tiny $\pm 1746$}	    \\ \cline{2-4} 
                                        &Semi-Bandit	 	&$0,859$ {\tiny $\pm 0,004$}    &$840$  {\tiny $\pm 912$}	        \\ \cline{2-4}  
                                        &P-BSB-RE	        &$0,840$ {\tiny $\pm 0,005$}    &$2476$ {\tiny $\pm 1359$}	    \\ \cline{2-4}  
                                        &P-BSB-OE	        &$0,836$ {\tiny $\pm 0,004$}    &$411$  {\tiny $\pm 265$}	        \\ \cline{2-4} 
	                                    &P-BSB-RD	        &$0,838$ {\tiny $\pm 0,002$}    &$1288$ {\tiny $\pm 1113$}	    \\ \hline \hline 

\multirow{5}{*}{TS}	        	        &Bandit		        &$0,825$ {\tiny $\pm 0,002$}    &$928$  {\tiny $\pm 1472$}	        \\ \cline{2-4}
                                        &Semi-Bandit	    &$0,857$ {\tiny $\pm 0,003$}    &$1938$ {\tiny $\pm 1511$}        \\ \cline{2-4}
                                        &P-BSB-RE	        &$0,845$ {\tiny $\pm 0,004$}    &$1206$ {\tiny $\pm 1211$}	    \\ \cline{2-4}
                                        &P-BSB-OE	        &$0,837$ {\tiny $\pm 0,003$}    &$545$  {\tiny $\pm 541$}	        \\ \cline{2-4}
	                                    &P-BSB-RD	        &$0,839$ {\tiny $\pm 0,007$}    &$1079$ {\tiny $\pm 1396$}        \\ \hline \hline 

\multirow{5}{*}{UCB}	        	    &Bandit	            &$0,832$ {\tiny $\pm 0,005$}    &$1171$ {\tiny $\pm 1304$}        \\ \cline{2-4}
                                        &Semi-Bandit	    &$0,842$ {\tiny $\pm 0,002$}	&$4163$ {\tiny $\pm 1512$}      \\ \cline{2-4} 
                                        &P-BSB-RE	        &$0,830$ {\tiny $\pm 0,004$}	&$1530$ {\tiny $\pm 1566$}      \\ \cline{2-4} 
                                        &P-BSB-OE	        &$0,823$ {\tiny $\pm 0,004$}	&$774$  {\tiny $\pm 905$}        \\ \cline{2-4}
	                                    &P-BSB-RD	        &$0,826$ {\tiny $\pm 0,002$}	&$1580$ {\tiny $\pm 1542$}	    \\ \hline \hline 

\multirow{5}{*}{UCB2}	        	    &Bandit	            &$0,796$ {\tiny $\pm 0,002$}	&$948$  {\tiny $\pm 1041$}	        \\ \cline{2-4}
                                        &Semi-Bandit	    &$0,796$ {\tiny $\pm 0,002$}	&$886$  {\tiny $\pm 950$}	        \\ \cline{2-4}
                                        &P-BSB-RE	        &$0,790$ {\tiny $\pm 0,002$}	&$1554$ {\tiny $\pm 1611$}	        \\ \cline{2-4}
                                        &P-BSB-OE           &$0,801$ {\tiny $\pm 0,003$}	&$889$  {\tiny $\pm 1213$}	        \\ \cline{2-4}
	                                    &P-BSB-RD	        &$0,792$ {\tiny $\pm 0,001$}	&$1734$ {\tiny $\pm 2235$}	        \\ \hline 

\end{tabular}
}
\caption{Non-Contextual case - $k=10$ - \textbf{Jester Dataset}} \label{table:Jester}
\end{table}

\begin{table}[ht]
  \centering
\renewcommand{\tablename}{Table}
\resizebox{0.5\textwidth}{!}{
\begin{tabular}{|c|c|c|c|} 
\cline{1- 4}
\textbf{Algorithm}	&\textbf{Strategy}	&\textbf{$Acc(T)$}	&\textbf{$t_c$}		 \\ \hline\hline  

\multirow{5}{*}{LinTS} 	                &Bandit	        &$0,996$ {\tiny $\pm 0,001$}	    &$330$ {\tiny $\pm 253$}	\\ \cline{2-4}
                                        &Semi-Bandit	&$0,995$ {\tiny $\pm 0,001$}	    &$732$ {\tiny $\pm 472$}	\\ \cline{2-4}
                                        &P-BSB-RE	    &$0,994$ {\tiny $\pm 0,001$}     &$627$ {\tiny $\pm 505$}	 \\ \cline{2-4}
                                        &P-BSB-OE	    &$0,994$ {\tiny $\pm 0,001$}	    &$391$ {\tiny $\pm 417$}	 \\ \cline{2-4}
	                                    &P-BSB-RD	    &$0,994$ {\tiny $\pm 0,001$}	    &$491$ {\tiny $\pm 330$}	 \\ \hline \hline

\multirow{5}{*}{LinUCB}         	    &Bandit	        &$0,994$ {\tiny $\pm 0,001$}	    &$944$ {\tiny $\pm 513$}	 \\ \cline{2-4}
                                        &Semi-Bandit	&$0,992$ {\tiny $\pm 0,001$}	    &$1582$ {\tiny $\pm 596$}  \\ \cline{2-4}
                                        &P-BSB-RE       &$0,991$ {\tiny $\pm 0,001$}	    &$1034$ {\tiny $\pm 541$}	 \\ \cline{2-4}
                                        &P-BSB-OE       &$0,990$ {\tiny $\pm 0,001$}	    &$644$ {\tiny $\pm 399$}	\\ \cline{2-4}
	                                    &P-BSB-RD	    &$0,990$ {\tiny $\pm 0,002$}	    &$1028$ {\tiny $\pm 662$}	    \\ \hline

\end{tabular}}
\caption{Contextual case - $k=10$ - \textbf{MoviesLens Dataset}}\label{table:Movies}

\end{table}
In order to evaluate the impact of each strategy on the agent's learning capacity independently from each \textit{COM-MAB / COM-CMAB} algorithm used, we compute and observe the average global accuracy and convergence time (See Table~\ref{table:resultats_moyens}), i.e., considering the results obtained by all algorithms applied with a given strategy on a specific dataset.

Thus, approaches are compared by evaluating the global accuracy average they obtain (See Table~\ref{table:resultats_moyens}) which is computed from results of Table~\ref{table:Jester} and Table~\ref{table:Movies} as follows:

     \begin{equation}\label{(4)}
    \hspace{-5px}
\forall \pi \in \Pi:
    M_{\phi}= \sum_{\pi=1}^{|\Pi|} \frac{Acc_{\phi}^{\pi}(T)}{|\Pi|}
    \end{equation}

Furthermore, approaches are compared with their convergence time averages. The latter is computed using a similar computation than for global accuracy (i.e., replace $Acc(T)$ by $t_c$ in Equation~\ref{(4)}). 


\subsubsection{\textbf{Results on Jester and MoviesLens}}

In Table~\ref{table:resultats_moyens}, we observe the global accuracy - $Acc(T)$ - and convergence time - $t_c$ - averages (See equation~\ref{(4)}) for each approach on \textbf{Jester} and \textbf{MoviesLens} datasets.

\begin{table}[ht]
  \centering
\renewcommand{\tablename}{Table}
\resizebox{0.5\textwidth}{!}{
\begin{tabular}{c|c|c|c|c|} 

\cline{2- 5}

\multicolumn{1}{c}{} & \multicolumn{2}{|c|}{\textbf{Jester}}  & \multicolumn{2}{|c|}{\textbf{MovieLens}}  
\\ \hline

\multicolumn{1}{|c|}{\textbf{Strategy}}	&$Acc(T)$ &$t_c$	&$Acc(T)$ &$t_c$		 \\ \hline\hline  

\multicolumn{1}{|c|}{Bandit}	    &$0,822$	    &$1127$	    &$0,995$	 &$637$     \\ \hline 
\multicolumn{1}{|c|}{Semi-Bandit}	&$0,839$	    &$1957$	    &$0,994$     &$1157$	\\ \hline 
\multicolumn{1}{|c|}{P-BSB-RE}	    &$0,826$	    &$1692$     &$0,993$     &$831$	    \\ \hline 
\multicolumn{1}{|c|}{P-BSB-OE}	    &$0,824$	    &$655$	    &$0,992$     &$517$	    \\ \hline 
\multicolumn{1}{|c|}{P-BSB-RD}	    &$0,824$	    &$1420$	    &$0,992$     &$759$	    \\ \hline

\end{tabular}}
\caption{Global Accuracy $Acc(T)$ and convergence time $(t_c)$ averages for each strategy}\label{table:resultats_moyens}
\end{table}

\subsubsection{\textbf{Statistical hypothesis testing}}

First, we compute a \textit{Kruskal-Wallis (KW)} test to highlight inequalities between the algorithms’ results (for each strategy) i.e., we test the null hypothesis $H_0$: \guillemotleft \textit{There is no significant difference between the algorithms’ results (medians)} \guillemotright.
Whenever the \textit{KW} test indicates that there are differences between the results, we compute \textit{Wilcoxon signed rank tests (RW)} over the global accuracy and convergence time results, i.e., we test the null hypothesis $H_0$: \guillemotleft \textit{There is no significant difference between the results of each pair of algorithms}\guillemotright.

In the following result analysis sub-section,  we will indicate for each case if the null hypothesis is rejected or accepted, and then give the corresponding \textit{p}-value.

\subsubsection{\textbf{Results analysis}}
\paragraph{Convergence time} Although we observe a slight edge of using \textbf{P-BSB-OE} compared to other methods in both contextual and non-contextual cases, there are no significant differences between them (\textit{KW} : $p > 0.05$). 

\paragraph{Global Accuracy}
\textit{KW} tests indicate that there are significant differences between global accuracy results of the 5 compared approaches, for each algorithm and in both contextual and non-contextual cases (\textit{KW}: $p < 0.01$).

In the non-contextual case, each setting of \textbf{P-BSB} eventually obtained a better accuracy compared to the \textbf{Bandit} approach (\textit{RW}: $ p < 0.01$). Nevertheless, \textbf{Semi-Bandit} strategy obtain better results compared to any other method (\textit{RW}: $ p < 0.01$). 
Moreover, \textbf{P-BSB-RD} and \textbf{P-BSB-OE} obtain similar accuracy results (\textit{RW} : $p > 0.05$).

In the contextual case, the \textbf{Bandit} approach outperforms any other methods (\textit{RW}: $ p < 0.01$).
Furthermore, there are no significant differences between \textbf{P-BSB-RE},  \textbf{P-BSB-RD} and \textbf{P-BSB-OE} approaches (\textit{RW}: $p > 0.05$).

\paragraph{Some observations}
In order to understand how much our proposal is cutting edge for several real-world application, we need to remind that our experimental results are obtained with \textbf{P-BSB} only observing 40\% of the feedbacks in the best cases (i.e., iterations where $\psi_t$ is maximal: 4) , and in the worst situations without observing any feedback (i.e., iterations where $\psi_t$ is minimal: 0). Hence, \textbf{P-BSB} approaches only use, on average 20\% of the feedbacks employed by the competitive methods. We strongly believe this will be an essential advantage when dealing online with real-world applications where users do not regularly accept to participate and give their feedbacks. 
The results we obtain on \textbf{RS-ASM}, \textbf{Poker Hand} and \textbf{Covertype} datasets are in line with those obtained by \textit{Jester} and \textit{Movie Lens} and confirm previously observed trends (with $k=3$ and $0\leq \psi_t  \leq3$). These results strengthens our confidence on the adequacy of our approach for recommendation systems built upon online learning techniques. 

\paragraph{\textbf{Conclusion}}
According to our experimental results, the main goal of \textbf{P-BSB},  i.e., ensuring a global accuracy similar to that of original competitive methods despite a restricted, or even occasionally null, number of feedbacks - is reached by all of the \textit{RE}, \textit{OE} and \textit{RD} settings.



\section{Conclusion}\label{Conclusion}

Our final goal is to integrate recommendation system driven by user needs in the special case of marine environment, where the rewards vector $R_t$ can be sparse due to partial user feedback. 

Thus, in this paper, we implemented and experimented a combinatorial setting on several Multi-Armed Bandit single-play algorithms.
We evaluated them in terms of global accuray and convergence time with various real-world datasets.
Our results tend to reinforce the idea that combinatorial approaches are relevant for recommender systems and similar applications.

The main contribution of our article is the proposition of a novel approach for considering user feedback: \textbf{P-BSB}.
In this article we experiment three settings of \textbf{P-BSB} : 1) \textbf{RE} which observes rewards associated to the $\psi_t$ arms with the best reward expectations ; 2) \textbf{OE} whose strategy is to pull the $\psi_t$ arms of $S_t$ for which rewards values have been the less frequently observed at iteration $t$; 3) \textbf{RD} which selects $\psi_t$ arms of $S_t$ at random.

In both contextual and non-contextual cases, the partial approach  combining \textit{Bandit} and \textit{Semi-Bandit} strategies, offers quite similar performances to original methods even though it uses a restricted number of user feedbacks.

The acquisition of user feedback and its efficient use remains a major challenge in the field of machine learning. Our results are thus promising for a further online evaluation on our real-world marine recommendation system.


Hence, as perspectives it seems relevant to extend our approach in order to enable it to dynamically select the most adapted strategy for considering users' feedback.


\section*{Acknowledgments}
This work has been carried out by the KARA TECHNOLOGY company in partnership with LERIA laboratory (University of Angers, France) and ESEO-TECH (Angers, France) and, with the support of the National Association for Research and Technology (ANRT, France).


\bibliographystyle{unsrt}  
\bibliography{references}  






\end{document}